\documentclass{article} 
\usepackage{amssymb} 
\usepackage{graphicx}
\usepackage{multirow} 
\usepackage[square,numbers]{natbib}
\usepackage{url} 
\usepackage{lscape} 
\usepackage{amsmath}
\usepackage{authblk}
\usepackage{rotating}
\usepackage{hyperref}

\makeatletter \def\url@leostyle{%
\@ifundefined{selectfont}{\def\UrlFont{\sf}}{\def\UrlFont{\small\ttfamily}}}
\makeatother
\begin{document} 
\title{Learning like humans \\with Deep Symbolic Networks} 
\author{Qunzhi Zhang\footnote{Email address: qzhang@ethz.ch}~ and Didier Sornette} 
\affil{ETH Zurich\\ Department of Management, Technology and Economics} 
\maketitle
\begin{abstract} We introduce the Deep Symbolic Network (DSN) model,
    which aims at becoming the white-box version of Deep Neural Networks
    (DNN). The DSN model provides a simple, universal yet powerful
    structure, similar to DNN, to represent any knowledge of the world,
    which is transparent to humans. The conjecture behind the DSN model
    is that any type of real world objects sharing enough common
    features are mapped into human brains as a symbol. Those symbols are
    connected by links, representing the composition, correlation,
    causality, or other relationships between them, forming a deep,
    hierarchical symbolic network structure. Powered by such a
    structure, the DSN model is expected to learn like humans, because
    of its unique characteristics. First, it is universal, using the
    same structure to store any knowledge. Second, it can learn symbols
    from the world and construct the deep symbolic networks
    automatically, by utilizing the fact that real world objects have
    been naturally separated by singularities.  Third, it is symbolic,
    with the capacity of performing causal deduction and generalization.
    Fourth, the symbols and the links between them are transparent to us,
    and thus we will know what it has learned or not - which is the key
    for the security of an AI system. Fifth, its transparency enables it
    to learn with relatively small data. Sixth, its knowledge can be
    accumulated. Last but not least, it is more friendly to unsupervised
    learning than DNN. We present the details of the model, the
    algorithm powering its automatic learning ability, and describe its
    usefulness in different use cases. The purpose of this paper is to
    generate broad interest to develop it within an open source project centered on the Deep Symbolic Network
(DSN) model towards the development of general AI.  
\end{abstract}

\pagebreak

\tableofcontents

\pagebreak

\section{Introduction} 
Deep learning\cite{LeCun_Bengio_Hinton_2015} has
achieved great success in many machine learning or artificial
intelligence (AI) areas, and it is attracting the attention of the
entire scientific community\cite{Appenzeller_2017}. In this paper, we
study the general AI problem from a statistical physics point of view,
and introduce a Deep Symbolic Networks (DSN) model to generalize the
Deep Neural Networks (DNN) model in Deep learning. The Deep Symbolic Networks
(DSN) model aims at solving the drawback of Deep
learning that the obtained models remain  black-boxes\cite{Castelvecchi_2016}.
In other words, DSN aims at  becoming a white-box version of
DNN. 

We study the problem from a different angle than DNN, by
naturally combining the DNN's structure with the methodologies (i) of 
Statistical Physics dealing with the micro-macro problem and (ii) of  
 complex system theory dealing with emergence and hierarchies \cite{Sornette_2006}.
 In a nutshell, these theories describe the world in a deep,
hierarchical structure. 
Indeed, physical matter constituting the Universe, i.e. anything occupying some space and 
having some mass, is made of elements. All these matter pieces are organised in many
composition layers, from the microscopic level to the macroscopic level.
From the micro to the macro levels, atoms make molecules, molecules
make cells, cells make organs and body parts, organs and body parts make
plants, animals (and human beings). Animals form communities, people
form societies, states, nations, unions... Soil particles
makes lands, water drops make rivers and oceans. Lands and rivers make
continents, continents and oceans make our Earth. The Sun, our Earth and
other planets make our solar system, billions of solar systems
make our galaxy, and billions of galaxies make the known universe. 
Similarly, from the macro to the micro levels, atoms have their own
hierarchical structures, formed of a nucleus and orbiting electrons,
the nucleus being formed of protons and neutrons, themselves
made of quarks, which may in their turn be conceived as vibrational modes
of fundamental strings... In a word,
the world in all its rich structure and complexity has deep, hierarchical structures at all scales,
though the natural laws gluing these structures together may be different
at different scales. 

While intelligence is nothing but about
learning and using this complex world, we conjecture that there is a
universal way of preserving knowledge at all these scales in human brains. More precisely, we
conjecture that the power of human brains benefits from a well-known tool in
statistical physics called coarse graining. The idea is that our brains
are good at clustering similar physical matter sharing enough common features at
any scales, and map them to the same conceptual symbol. In this way,
humans can significantly reduce the various sources of noise in the information produced by the real world. It is also
an efficient way in the economic sense, as learning from noise will not increase
humans' utility functions, but only the consumption of more energy.

\section{Definition of Deep Symbolic Networks}

\subsection{Recursive hierarchical model}

Let us call some specific coarse-grained physical matter at
some scale an ``object'', and denote it by $f_{n,i}(x)$, where $n$ is its
hierarchical layer, $i$ is its index that allows us to identify it on that layer, and $x$ is
its domain, for instance, the space it occupies. The function $f_{n,i}$
approximates thus the details at each point in the space the object
occupies. The level of coarse-graining is determined by a trade-off between
the costs and benefits of learning. To be specific, let us consider 
a physical thing to be coarse-grained, perceived as 
the object plus an error term, denoted by $\epsilon_{f_{n,i}}$, which can be
modelled as a Gaussian centred random variable.
We have therefore a simple mathematical formula to describe physical matter
\begin{equation}\label{matter}
    \mathbf{matter}:=\mathbf{f_{n,i}(x)}+\mathbf{\epsilon_{f_{n,i}}}
\end{equation}
With these notations, we can formulate the core of 
the Deep Symbolic Network (DSN) model as follows:
\begin{equation}\label{dsn}
    \mathbf{f_{n,i}(x)}+\mathbf{\epsilon_{f_{n,i}}}=\mathbf{\sum_{j=1}^{k_{n,
    i}}(a_{n_j, j} +
    \epsilon_{a_{n_j,
    j}})\left[f_{n_j,j}\left(\frac{x_j - (b_{n_j, j} + \epsilon_{b_{n_j,
    j}})}{c_{n_j, j} + \epsilon_{c_{n_j, j}}}\right)+ \epsilon_{f_{n,j}}\right]}
\end{equation}
The recursive model simply states that matter at layer $n$ is made of
matter at other layers, $n_j$, where $j=1...k_{n,i}$, and $k_{n,i}$ is
the total number of the direct composition objects of the object $f_{n,i}(x)$. 
The recursive model presents
the fact that we observe from the real world, and 
there is no loop in the model, i.e. no part of an object is
made of itself. The parameters $a_{n_j,j}$, $b_{n_j,j}$, and $c_{n_j,j}$
represent the states of the objects, such as magnitude, place, size, and so on.
Changes in the values of these parameters correspond to operations on
the objects such as amplification in magnitude, displacement in space, or
variation in size. There could also be other linear operations, for instance,
rotation, which we do not discuss here for the sake of simplicity.

For some objects, the state parameters such as 
$a_{n_j,j}$, $b_{n_j,j}$, $c_{n_j,j}$, can have certain constraints. For instance, the
natural languages are constrained by the grammars. Such constraints can
be expressed mathematically as following:
\begin{equation}\label{constraint}
    \mathbf{C(a_{n_1,1}, b_{n_1,1}, c_{n_1,1}, ..., a_{n_j,j},
    b_{n_j,j}, c_{n_j,j}) > \theta}~,
\end{equation}
where $\theta$ denotes some threshold.

\subsection{Identifying operators of symbols}

By assuming that each object $f_{n,i}(x)$ is mapped in human mind to a
symbol, denoted by $S_{n,i}$, equation~(\ref{dsn}) becomes our Deep
Symbolic Networks (DSN) model. Formally,
\begin{equation}
    \mathbf{S_{n,i}}:=\mathbf{f_{n,i}(x)}
\end{equation}
In this way, all physical matter or abstract concepts are symbolized. The atom,
the molecule, the cell, the grain of sand, the water droplet, the river, the land, the world, the
universe, the number, the space, and so on and so forth, are symbols. As a
matter of fact, humans have already symbolized these symbols in natural languages.

The goal of the symbolization is to represent a rich variety of physical matter or
abstract concepts sharing common features with a single abstract symbol. To
this end, it is crucial that a symbol $S_{n,i}$ must be equipped with at least one
identifying operator so that it can identify physical matter or abstract
concepts. When an identifying operator of it is applied to data of certain physical
matter or abstract concept, it will determine if the physical matter or abstract
concept can be represented by the symbol, i.e., belongs to the class of things
represented by the symbol. 

We assume that an identifying operator is a linear operator, denoted by $M_{n,i}$.
On the one hand, if it is applied to data of certain physical matter or abstract
concept picked from the class of things the symbol represents, the
output should exceed a threshold.
On the other hand, if the physical matter or abstract concept is irrelevant, the
output should be below the threshold. Mathematically, this can be expressed as follows.
We call a linear operator $M_{n,i}$ an
identifying operator of the symbol $S_{n,i}$ if it satisfied
\begin{equation}\label{id}
    \mathbf{M_{n,i}(f'_{n,i}) > v > M_{n,i}(f'_{m,j})}, \forall{m\neq n,
    j\neq i}~,
\end{equation}
where $f'_{n,i}$ denotes that the symbol is normalized, and $v$ is the
decision making threshold.

A symbol can have more than one identifying operator, and the final decision in
that case is the output of a Boolean combination of the
decisions of all identifying operators. For instance, the symbols ``pests'' and
``beneficial insects'' must be identified first to be animals or insects, and
then to be harmful or to be beneficial. An identifying operator defines a
symbol, a Boolean combination of identifying operators of many different
symbols generates a new identifying operator, thus defining a new symbol. Rich
symbols can be built in this way.

A symbol can be given a name or just be anonymous, and it can be formed in a
deep, hierarchical way as depicted in expression (\ref{dsn}). Indeed, the symbols, and
the links between the symbols in such a deep, hierarchical structure provide a universal
way of storing any knowledge about the physical world and any knowledge
abstracted from the physical world.

\section{Structure and properties of Deep Symbolic Networks}

\subsection{Links between the symbols}

\begin{sidewaysfigure}
    \caption{An illustration of the Deep Symbolic Networks (DSN). The pictures are
    taken from the Wikipedia website\cite{Wikipedia}.}
    \centering
    \includegraphics[width=\textwidth]{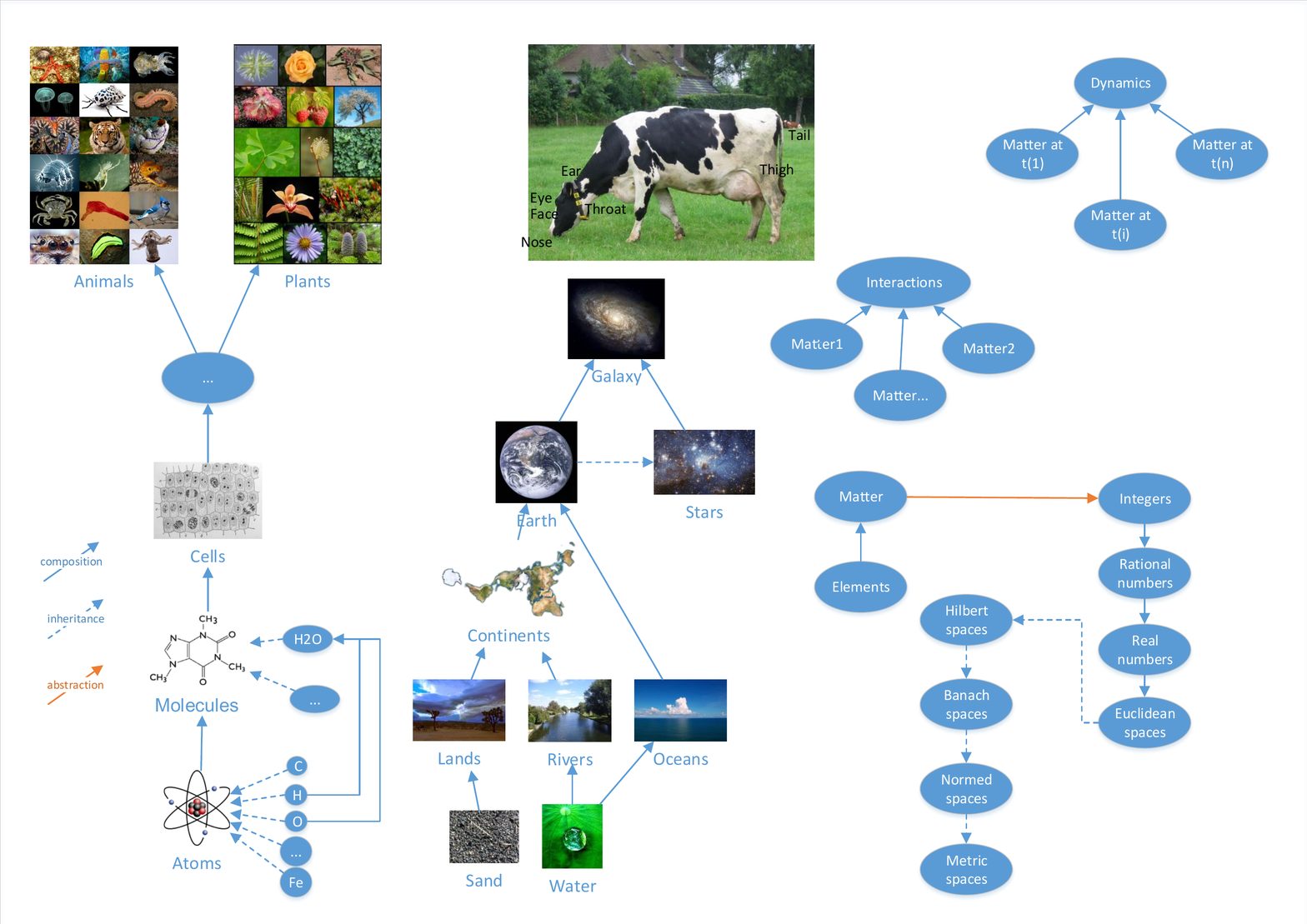}
    \label{fig:dsn}
\end{sidewaysfigure}

Suppose we have already constructed a deep symbolic network as in
model~(\ref{dsn}), with all symbols equipped with their identifying operators. One
notices immediately the composition links between the symbols, emerging from the
relationships between them.
Figure~\ref{fig:dsn} is an illustration of the Deep Symbolic Networks (DSN) model, including
many well-known symbols, and their links representing what makes what.
On the left side of the figure, how physical matter composes the universe is represented in
the same symbolic form.

\subsubsection{Links by the composition relationship}
The composition relationship is easy to understand, just indicating what
kinds of physical matter or abstract concepts compose other kinds. It links
many symbols together, such as the atom and the molecule, the molecule and the
cell, the cell and the organ and many other body parts, to name only a few.
A concrete example is 
the $\mathbf{H_2O}$ molecule, which is made of
two Hydrogen atoms and one Oxygen atom. The relationship is about that the Hydrogen atom and the Oxygen
atom compose the $\mathbf{H_2O}$ molecule, and that the $\mathbf{H_2O}$ molecule is
composed by the Hydrogen atom and the Oxygen atom. The example shows therefore the
composition relationship is two-way. Figure~\ref{fig:dsn} depicts the
composition relationship only one-way, as one direction of the relationship
implies its counterpart automatically. This is true for all relationships, therefore in the figure,
we present all relationships one-way.

\subsubsection{Links by the inheritance relationship}
The links by the inheritance relationship,
however, are not included  explicitly in model~(\ref{dsn}). This relationship indicates
that one kind of physical matter or abstract concepts belongs to another broader kind. For instance, hens,
cows, and horses all belong to the cattle kind, which itself belongs to the animal
kind. In this relationship, the animal symbol is at a lower layer, the cattle
symbol is above it, and the hen, cow, and horse symbols are above them. It is
clear that the lower layer symbols capture more general common features of the
physical matter or abstract concepts they represent, while
the upper layer symbols capture more specific common features. A similar idea
exists in object oriented programming\cite{Rentsch_1982}.

This relationship is again two-way, which are generalization and concretization respectively.

When it comes to the identifying operators, the higher level, more concrete symbols can be identified by the identifying
operators of its lower level, more generalized symbols, as long as they are
connected by this relationship. It thus provides a way to look for more
general symbols from identifying operators directly; i.e., we look for
identifying operators whose inputs are identifying operators.

\subsubsection{Links by dependence}
This relationship indicates that the symbols tend to link to the same symbol,
either implicitly or explicitly. For instance, orange and apple are different,
but they are related as they both belong to the fruit kind. A computer monitor
and a mouse are also related because they are both a part of a computer system.

\subsubsection{Links by causality}
According to Wikipedia, ``Causality (also referred to as causation, or cause
and effect) is the natural or worldly agency or efficacy that connects one
process (the cause) with another process or state (the effect), where the first
is partly responsible for the second, and the second is partly dependent on the
first. In general, a process has many causes, which are said to be causal
factors for it, and all lie in its past. An effect can in turn be a cause of, or
causal factor for, many other effects, which all lie in its future.'' A process
or a state is represented by a symbol in the Deep Symbolic Networks (DSN) model,
and so does its effect. The causal-effect symbolic group composes a new symbol.
The causality relationship is thus a special form of the composition
relationship, and the involved symbols are often related to time, because of the
interests of humans to predict the future.

\subsubsection{Links by abstraction}
On the right side of figure~\ref{fig:dsn}, a few mathematical
symbols are represented. These symbols emerge also from model~(\ref{dsn}). For instance, the
integer number emerges from the parameter $k_{n,i}$ in model~(\ref{dsn}), i.e.
the number of the direct underlying symbols that compose the symbol $f_{n,i}$. A parent
symbol composed by a list of identical child symbols can be abstracted to 
a combination of the child symbol and the number symbol, indicating the number of
children in the composition relationship. From the state
parameters $a_{n_j,j}$, $b_{n_j,j}$, and $c_{n_j,j}$, one can derive conceptual
symbols such as directions, positions, and sizes.

\subsubsection{Higher order links}
A symbol can connect to another one implicitly by a path of many links, along the
network structure of the Deep Symbolic Networks (DSN) model. Those two
implicitly connected symbols can be connected directly with a link composed by
all links in the middle, namely a higher order link. 

\subsubsection{Links can also be represented by symbols}
A concrete link between two symbols can also be seen as a symbol. It is composed by
the involved symbols, as in the following set: 
\begin{equation}
    \{\mathbf{S_1}, \mathbf{S_2}\}~,
\end{equation}
if and only if the symbol $S_1$ is composed with symbol $S_2$. A deep
symbolic network has to store all these links. A conceptual composition link
symbol can be defined as an aggregation of all concrete composition links; i.e,
its identifying operator can identify all the concrete composition links. 

\subsection{Dynamics, interactions and processes}
The upper right side of figure~\ref{fig:dsn} depicts the symbols
associated with dynamics and interactions. A process is usually a series of interactions producing 
final outputs through the dynamics of the interactions. They often involve time,
as they happen in specific order.

\subsubsection{Dynamics}
Dynamics can be represented by a symbol with its states changing with
time. According to model~(\ref{dsn}), a symbol is defined by its underlying
components and the state parameters such as $a_{n_j,j}$, $b_{n_j,j}$, and
$c_{n_j,j}$. The dynamics symbol represents thus its composing symbol evolving in the
time dimension. A specific type of dynamics can thus be identified by similar
changes of the state parameters in time. 

\begin{figure}
    \caption{An illustration of a symbol representing dynamics. A thing is falling to the
    ground, where $t1$, $t2$, $t3$, $t4$ denote the elapsing of the time.}
    \centering
    \includegraphics[width=1.2\textwidth]{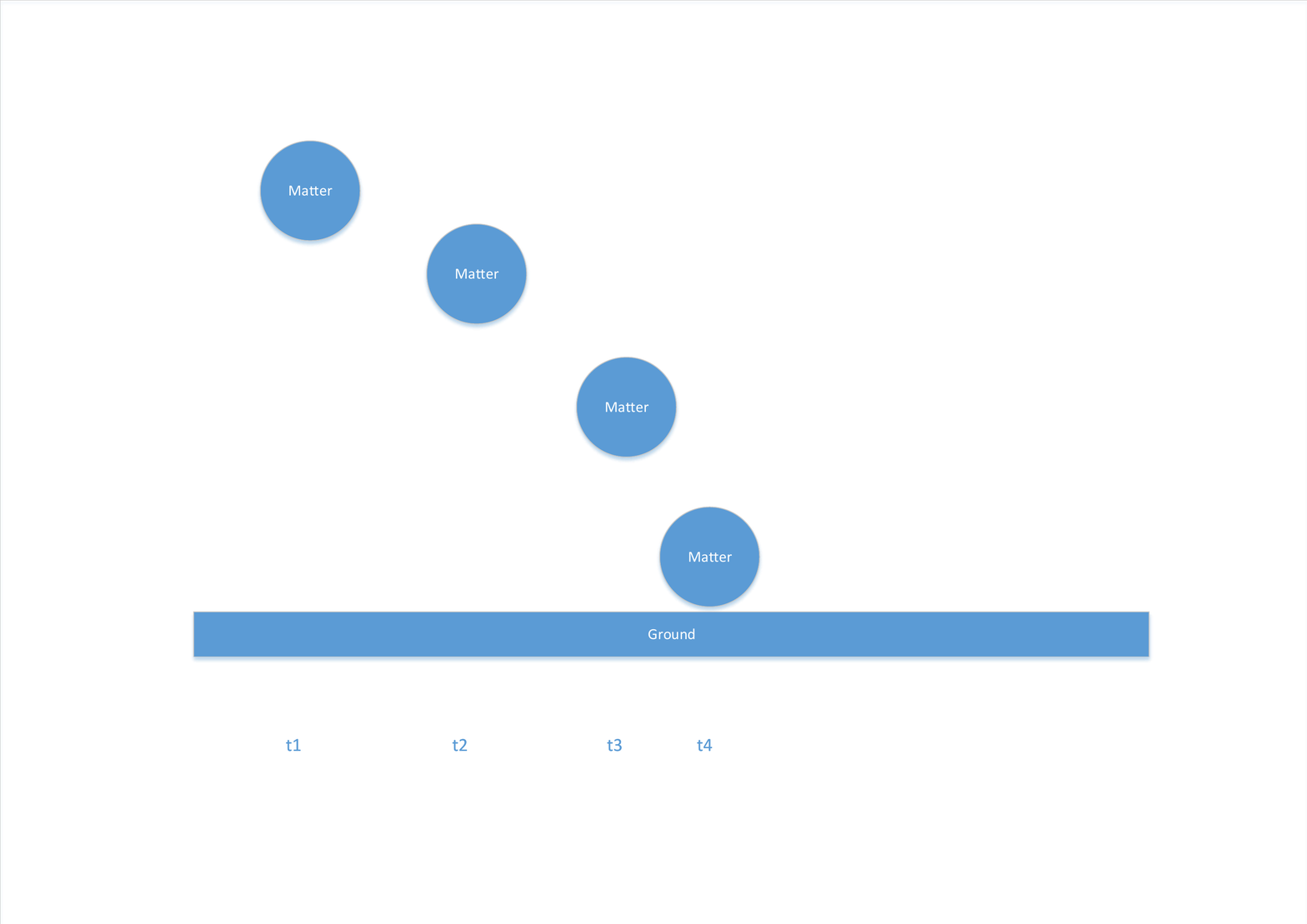}
    \label{fig:falling}
\end{figure}

Consider the concrete example of the dynamics symbol representing `falling', which
describes the dynamics of an object falling to the ground. The falling symbol involves
thus two symbols, certain physical matter, and the ground. The combination of
the physical matter and the ground forms the underlying symbol of the falling
symbol, and the falling symbol is about how the distance between the physical
matter and the ground change along time, as depicted in the
figure~\ref{fig:falling}.

\subsubsection{Interactions}
All human actions generate interactions between humans and physical matter or
other humans. An interaction has its participants, its outcomes, and is often
linked with some dynamics. Humans change the world through interactions with the
world. Interactions between physical matter create the colorful world. The same
action, mixing two different things, can take place in both the real world and
the virtual world. For instance, we can mix chocolate and milk together to
produce chocolate milk, or we can mix a character with a number to generate
another symbol, such as ``number1''. The two examples are not so
different from each other in a deep symbolic network. The former one has to be
carried out in the physical world, and the latter one has to be done by a
function in the computer world. Nevertheless, they are generalized to be the
same interaction in a deep symbolic network.

\subsubsection{Processes}
The operation of a machine is a process to convert one thing to another one. A
process is nothing but a series of interactions between physical matter or
conceptual ideas, such as the process of making a car, or the process of proving
a mathematical theorem. In a deep symbolic network, they can be both
generalized to the same symbol.

\subsection{Storing the Deep Symbolic Networks (DSN) to storage media}
The core ingredients of a deep symbolic network are symbols and the links
between them. Inside a symbol, we should store pointers to its direct underlying
symbols, and also pointers to the symbols it links to. A symbol is defined by
its identifying operators, which must also be stored. We shall see in the later
sections that the identifying operators are just vectors, or tensors with higher
dimensions. Along with the identifying operators, the thresholds in equation~(\ref{id})
must also be stored. One often needs to represent certain concrete physical
matter by a
symbol, for instance, the computer in an office. It is an instance of the
computer symbol. By borrowing terms from the object
oriented programming field\cite{Rentsch_1982}, we can think of the symbols as
classes, and their instances as objects. In the classes, we record only the
statistical facts of the state parameters $a_{n_j,j}$, $b_{n_j,j}$, and
$c_{n_j,j}$, while in their instances we shall record the concrete values. In
addition, the prior and posterior distributions of the deep symbolic network
learned from data must also be stored for future Bayesian decision making, as
discussed below in the section~\ref{genpr}.  The
solution of this kind of storage problems exists already in computer science, so
we stop discussing it any further.

It is obvious that the more symbols, links and instances of the symbols stored
in a deep symbolic network, the more powerful it is. What to store or how much
to store is a trade-off between
the benefits and the storage costs.

\subsection{Use of the Deep Symbolic Networks (DSN)}
\subsubsection{General principles of statistical inference and object generating
}\label{genpr}
We list a few practical tasks a deep symbolic network will face in the real
world, and introduce the general principles of accomplishing the tasks.
The practical tasks include cognitive tasks, decision making and the generative problem.
We examine them in turn.

{\bf Cognitive tasks}. A deep symbolic network must be able to map
physical matter or abstract concepts to symbols in its vocabulary correctly.
Naturally it can be done in a bottom up manner. For instance, if the input data
is a picture, the DSN will scan all the parts of the picture from a small scale to
the largest scale, and identify the first layer symbols, i.e. ground symbols at all
scales. That is done by iterating over the ground symbols and applying their
identifying operations. In this process, the prior distribution of each ground
symbols, i.e. $\mathbf{P}(f_{0,i})$, where $\mathbf{P}$ denotes the probability,
and $f_{0,i}$ denotes the $i$-th ground symbol, can be used to determine the
priority of scanning the ground symbols. When ground symbols are identified, it will then look
for higher level symbols using the posterior probabilities
$\mathbf{P}(f_{n,i}|f_{0,1},...,f_{0,k})$, where $f_{n,i}$ denotes a higher level
symbol than the ground level, which is an existing symbol in the network. The probabilities
$\mathbf{P}(f_{0,j}|f_{n,i})$, where $f_{0,j}$ is a ground symbol existing in
the data but not identified yet, can be used later to identify more details from
the ground level. Continuing this process, it will accomplish the cognitive task
and create a small network containing instances of identified symbols for
further information processing.

{\bf Decision making} mainly involves searching the paths between
the links and dealing with missing symbols in the deep symbolic network. Denote the input symbols it has
identified from the data by $f_{n_i, i}$, where $i$ is the index of each symbol,
and denote by $f_{m,j}$ a list of possible decisions to make, again represented by
symbols. DSN is thus finding the
$\mathbf{\arg\max_{j^*}}\mathbf{P}(f_{m,j^*}|f_{n_1,1},...,f_{n_k,k})$, a
Bayesian decision making problem, which can be done by using the following
formula:
\begin{equation}
    \mathbf{P}(f_{m,j}|f_{n_1,1},...,f_{n_k,k})=\sum_{f_{l_i,i}}\mathbf{P}(f_{m,j}|f_{l_1,1})\mathbf{P}(f_{l_1,
    1}|f_{l_2,2})...\mathbf{P}(f_{l_i,i}|f_{n_1,1},...,f_{n_k,k})~,
\end{equation}
as long as it can find the corresponding link path to realize the calculation.
A simple example is to answer the question ``From which direction does the Sun
rise?''. A deep symbolic network will identify the symbols such as the Sun,
rise, and direction, and it will find the Sun symbol, which linked to the rise
symbol, which in turn contains the direction symbol with a value ``east'', the
correct answer. Here, the rise symbol of the Sun is an instance of the more
general rise concept.

{\bf The generative problem} consists for example in drawing a picture, or writing
an article. This task can be done with model~(\ref{dsn}). As we can expect, to
generate better results, the deep symbolic network needs to learn harder for
more details on each layer.

The following sections show more concrete examples.

\subsubsection{Deep Symbolic Networks (DSN) for survival}
Suppose there is an existing deep symbolic network aggregating all the knowledge the
human kind needs, and let us imagine how it generates human-level intelligence with
the following thought experiment. 
Say an ancient hunter equipped with such a deep
symbolic network, encountered a wild cow. He would immediately receive visual
data input of the wild cow from his eyes, and start to process the
information. He would apply the symbol identifying operators 
to identify the ears, eyes, face, legs, and all other body parts of the wild cow,
as depicted on the upper middle side of the figure~\ref{fig:dsn}, and identify
the highest layer symbol, the wild cow, under the circumstance. He would then run
through the links of the wild cow symbol to check its usefulness, and he would
realize that it can be turned to food by hunting it. A path of symbol links
tells him how to turn the wild cow into his food, which is a series of
interactions between him and the wild cow. He therefore would apply the hunting
algorithm, represented by a hunting process, to the wild cow. In reality it is
not always possible to identify certain symbols deterministically due to lack of
information, for instance, a missing underlying symbol, so
he would have to make decisions based on the Bayesian inference as discussed in
the section~\ref{genpr}. To that end, the ancient hunter had to keep receiving
information and react to the new situations adaptively according to the
knowledge stored in his deep symbolic network.

\subsubsection{Automatic coding}
Automatic coding is a challenging task. The code needs to be first represented in
symbols in a deep symbolic network, and then translated to certain programming
languages to be executed by computers. The machine to which the task is assigned
must first identify the key symbols from the inputs and outputs of the task, and
then find a link path in its deep symbolic network to convert the inputs to the outputs.
To this end, it must have accumulated a huge amount of
knowledge represented in symbols and links. The following is a simple and
concrete example.

Suppose a machine is given a C++ coding task to read the content from a specific CSV file, sum
up the values in each column, and print out the results. The machine will start from
identifying the symbols, such as the file, the file content, the column, the
sum, the print, and so on. It will instantiate the involved symbols, in the same
way as instantiating objects according to their classes in C++ programming. The
instantiated symbols will be naturally linked with many other symbols in the
deep symbolic network, and many links are important for the task, for instance,
the file content is made of lines, a line is made of characters and a line
break, and the characters in a line contain data fields separating by certain
separator, normally `,'. The same data fields in the
lines make columns, and the data fields are strings, which must be converted to
numbers. The machine will follow these links to reach the output symbol, and print
the sum of the data fields. The path defines a process to convert the input to
the output, including interaction symbols, such as opening files, reading lines,
splitting data fields, converting strings to numbers, summing up numbers, and so
on. Translated to C++, it will be operations such as instantiating objects,
calling their methods to generate new objects, and so on and so forth, according to the definition of
the interaction symbols, until it
reaches the output. More specifically, entity symbols will be translated to C++ classes,
interaction symbols will be translated to class methods, and symbol composition
will be translated to class initialization. In the main function, the classes
will be instantiated, and the process will be started.

The automatic coding procedure often takes many iterations, 
due to uncertainty in how to interpret the inputs. For instance the
machine may assume that the data fields are separated by `,', according to the
posterior distribution of the separator symbol, but
it might actually be `:'. In that case it
has to open the file with a text editor and identify the data field separator 
visually, instantiating a new separator symbol to replace the old one, and
re-program the code accordingly.

Translating the symbolic process to C++ code needs good knowledge
about the C++ programming language. The machine needs to know all the library
interfaces, classes, methods, functions, and the C++ language specification as
well, and learn symbols and links from
them, so that it can translate all pieces of the symbol process into C++ clauses.
If a function computing outputs from certain inputs does not exist yet, it must find
a computer algorithm to implement it according to the grammar of the C++
language. It will not be able to accomplish the task even if there is a tiny
missing piece in its deep symbolic network.
 
In summary, to accomplish automatic coding, the machine has to accumulate a huge
amount of knowledge. Nevertheless, the brighter side is that, powered by a deep
symbolic network and an unsupervised learning method, a machine could be
able to scan all open source software and learn unsupervised all the symbols
and links, and thus acquire the knowledge to translate software from one
programming language to any other ones, as humans do, but it would do it in a much faster way.

\section{Learning of the Deep Symbolic Networks (DSN) model from data}
The Deep Symbolic Networks (DSN) model would have little value if it was not
able to learn from data, which are often visual data represented by pixels,
acoustic data represented by sound waves, or text encoded in natural languages.
Here, we base our discussions on visual data, but the ideas
and methods, because of their
universality, can be easily applied to other types of data. 

\subsection{Identifying operators}\label{idop}
\begin{figure}
    \caption{Examples of singularities separating different instances of
        different symbols.}
    \centering
    \includegraphics[width=\textwidth]{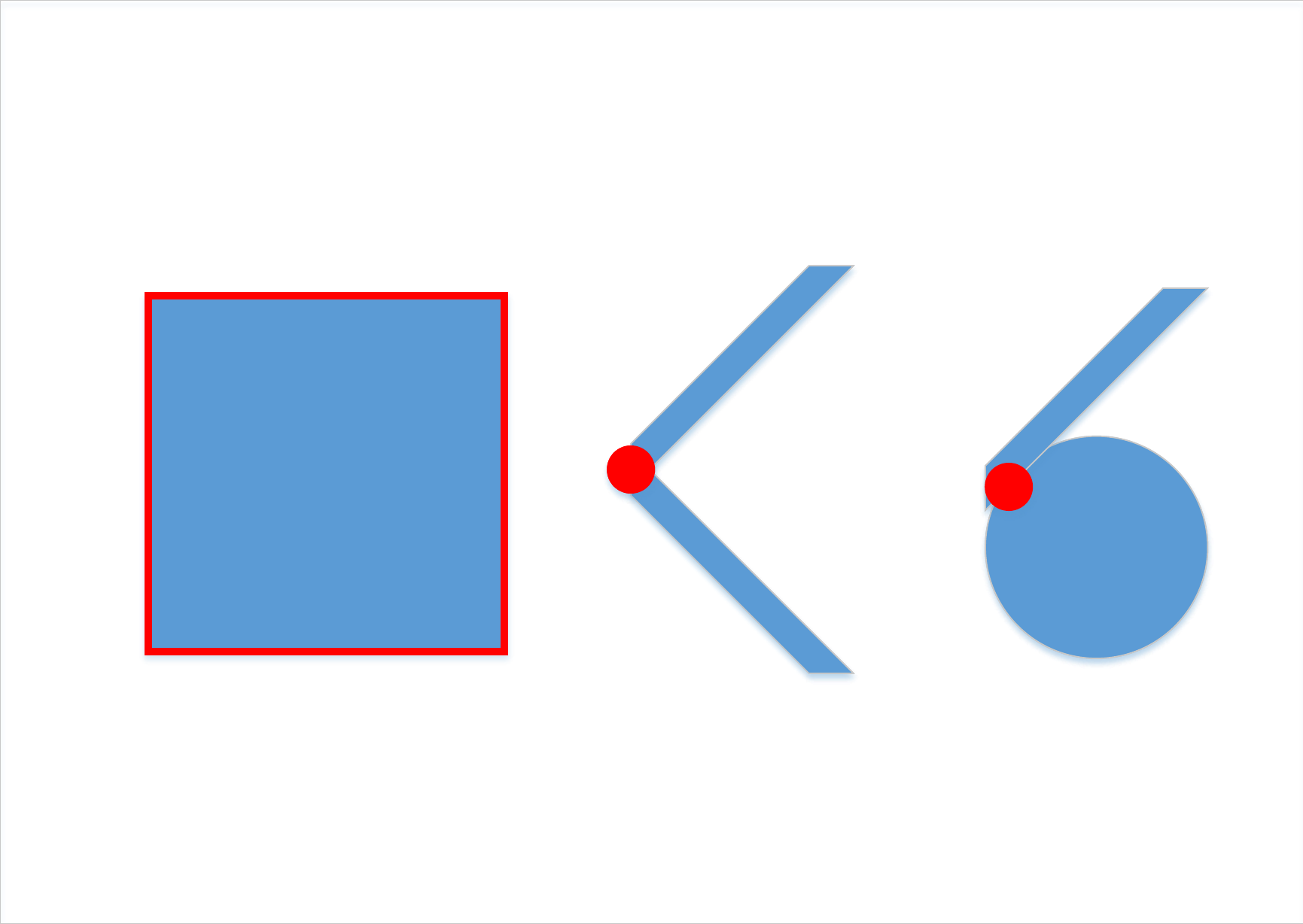}
    \label{fig:sig}
\end{figure}

Because of the multi-layer noises present in physical matter, it is usually
difficult to isolate symbols from one another, not to mention to identify them.
Nonetheless, symbols can be naturally isolated by singularities. Indeed,
as shown in model~(\ref{dsn}), if the function $f_{n,i}$ and its derivatives
are continuous everywhere in its domain $x$, it will be a whole integrated piece, without its
underlying building blocks. For visual data, the singularities are present in both
the color space and in the edges. In a nutshell, singularities are present on the
boundaries of the underlying building blocks of any symbol, separating the
building blocks naturally.   
Figure~\ref{fig:sig} shows three instances of singularities. In the first case,
the singularities are the red edges, because of the discontinuity in the color
space. In the second one, the singularity is a red connecting point,
because of the discontinuity in the slopes of the two lines it connects. In the third one,
the singularity is the red tangent point where the straight line touches the circle,
because of the discontinuity in the curvatures. The methods of identifying singularities
have been actively studied, and many well-known ones are based on wavelets 
analysis\cite{Mallat_2009}. 

Identifying one dimensional singularities on edges is more or less special, as
the two dimensional visual data have to be converted to one dimensional. 

With the singularities, we can isolate the symbols at all scales by looking for
bounding boxes of the symbols, which are rectangle areas isolating
approximately the symbols. For natural languages, it is even better that the
symbols are already presented in the dictionaries.

The ground symbols, i.e. the first layer symbols, must be among those isolated
ones. They form the bases of the symbol space, so their identifying operators
are themselves normalized. Indeed,  
given two different types of data pieces, $f_{n,i}(x)+\epsilon_{f_{n, i}}$ and
$f_{m,j}(x)+\epsilon_{f_{m, j}}$, containing two different symbols
$S_{n,i}:=f_{n,i}(x)$ and $S_{m,j}:=f_{m,j}(x)$, we need an identifying operator
$M_{n,i}$ to separate them according to equation~(\ref{id}). 
The operations of $M_{n,i}$ can be
represented by 
$M_{n,i}(f_{n,i})=g_{n,i}\cdot f'_{n,i}$, and $M_{n,i}(f_{m,j})=g_{n,i}\cdot
f'_{m,j}$, where $\cdot$ denote the dot product, $f'$ is the normalized $f$, and $g_{n,i}$ is another
normalized function represents $M_{n,i}$. 
So the identifying operator returns the cosine similarity between the
$g_{n,i}$ and the symbols. $f'_{n,i}$ is thus the best choice for $g_{n,i}$,
and it is true for all other ground symbols. As $f_{n,i}$ is represented by a
vector, so does its identifying operator.

Higher level symbols must be represented by their direct underlying building
blocks. Say a symbol $S_{n,i}$ is made of $S_{n_k,k}$, where $k=1,2,...$, it can
be represented by $(S_{n_1,1}, S_{n_2,2}, ...)$, i.e. a symbol vector. Its
identifying operator is again itself. The symbols are the orthogonal bases of
the symbol space, so the cosine similarity indicates the square of the fraction
of the same
compositing symbols. The state parameters of the direct underlying symbols, such as
$a_{n_j,j}$, $b_{n_j,j}$, and
$c_{n_j,j}$ in the model~(\ref{dsn}), store the state information of the symbol,
such as the size, place, color, and so on, so there must be another identifying
operator to operate on these state parameters. For instance, the digits 6 and 9
have similar underlying building blocks, while they appear always in different
places in different symbols. By applying an identifying operator to the state
parameters, we will be able to distinguish each from one another.

Because of the universal representative form of identifying operators, we can
look for identifying operators on identifying operators, thus defining more
general symbols.

\subsection{Unsupervised learning}\label{uns}

Unsupervised learning from data is a process of clustering similar things into
symbol categories layer by layer, until no new symbol can be identified. If the
learning is cumulative, the first step of the unsupervised learning is to
identify the known symbols from training samples, and then the next steps are no different
than the following algorithm of learning from scratch.

\begin{enumerate}
    \item Detect singularities at all data points at all scales in all training
        samples.
    \item Isolate the symbols by the singularity bounding boxes in all training
        samples, and cut down the data pieces from the singularity bound boxes.
        The symbol identifying will be always running on these data pieces. Of
        course we should record the hierarchical composition relationships
        between the data pieces.
    \item Iterate over all data pieces to cluster them, and thus find the
        ground symbols. We create an empty cluster list in the beginning.   
        When a data piece is picked, we create a new cluster, add the data piece
        into the cluster, and insert the cluster into the
        cluster list, if the cluster list is empty, and generate the identifying operator of this new
        cluster, which is the normalized data piece. If the cluster list is not
        empty, we apply the identifying operators of all clusters to the data
        piece. And we also rescale, rotate or transform linearly in other ways
        the data piece to get the best cosine similarity between the data piece
        and the existing clusters. If a result is higher than a threshold
        $\lambda_0$, it will be added into the corresponding cluster; if it cannot be added to
        any existing cluster, a new cluster will be created for it. If two
        different results are qualified, either the two clusters should be
        merged, or the threshold should be increased. When a new data piece is
        added into an existing cluster, its identifying operator should be
        updated by averaging all the data pieces in the cluster to reduce
        noises, including the new one. In this way, every data
        piece will be clustered in a cluster. We filter out those clusters with
        too few data pieces with a threshold $\mu_0$. The remaining clusters are the
        ground symbols we look for, defined by the identifying operators.\label{it:it}
    \item In this step, we go to higher layers. On each layer, there are again two
        thresholds, $\lambda_i$ and $\mu_i$, where $i$ is the number of the layer. Different than in the step~\ref{it:it},
        the data pieces are represented by the known symbols, as discussed in
        section~\ref{idop}. Here, the insight is to decrease the multi-layer noise.
        Specifically, when a noisy data piece is identified by a
        symbol, we remove all noises it carries. Another difference between this
        step and the step~\ref{it:it} is that we also look for the identifying
        operators for the state parameters, as discussed in
        section~\ref{idop}. The clustering method for the symbols and the state parameters are
        the same as in step~\ref{it:it}. We continue this step until no new
        symbols can be identified.\label{it:sym}
    \item We add links between the symbols by their composition relationship.
    \item We record the prior and posterior probabilities for the symbols.
\end{enumerate}

The basic assumption underlying this unsupervised learning algorithm is that the
symbols can be linearly separated, and the amplitudes of the noises presented in the data are small
enough. The size of the training samples depends on the number of symbols to
learn. The algorithm will get too many unnecessary symbols if the thresholds $\lambda_i$ and
$\mu_i$, where $i=0,1,...$, are too tight, or get too few symbols if they are
too loose. The algorithm has to know the approximate number of symbols to learn, or
find an area where the number of learned symbols are insensitive to the change
of the thresholds. In any case, the algorithm has to run a number of experiments
to learn better.

\subsection{Supervised learning}
There are much more tools available in supervised learning than in
unsupervised learning. One method is that we use the same unsupervised learning
algorithm in the last section, while find the best thresholds $\lambda_i$ and
$\mu_i$ by optimizing the objective function.

Another way is to learn the generative function in model~(\ref{dsn}) directly,
using the method similar to that developed in Ref.\cite{Patel_Nguyen_Baraniuk_2016}.

\section{Experiment}
\begin{figure}
    \caption{Some digits examples in the training samples of the MNIST database.}
    \centering
    \includegraphics[width=\textwidth]{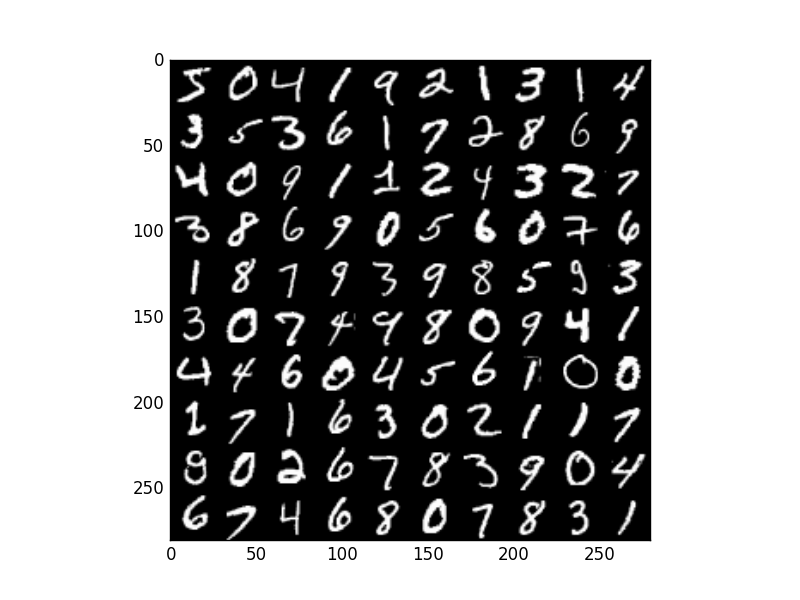}
    \label{fig:mnist}
\end{figure}

We now propose an experiment protocol to validate the Deep Symbolic Networks (DSN) model
~(\ref{dsn}) with the MNIST database of handwritten
digits\cite{Lecun_Bottou_Bengio_Haffner_1998}. Figure~\ref{fig:mnist} shows some
examples of digits in the training set of the MNIST database. The validation should
include both the unsupervised learning algorithm and the supervised learning
algorithm. The relationship between the success rates of the unsupervised
learning and the training sample sizes should also be studied. Moreover, we
should test if there are stable areas of the thresholds $\lambda_i$ and $\mu_i$,
so that the learned number of symbols is stable in the threshold range. 
as discussed in section~\ref{uns}

Digits share some basic building blocks, which can be inferred from
figure~\ref{fig:mnist}. A few obvious ones are straight
lines (in 1, 4, 5, 6, 7, 9), open curves (in 2, 3, 5), close curves, i.e.
circles (in 0, 6, 8 and 9), and connecting points or crossing points as well (in
all digits except 0 and 1). In addition, the open curves may comprise other more
fundamental building blocks. For instance, two open curves connected together
make 3, while each open curve is made of two curving segments connected by a
point. When one draws a closed circle to represent 0, one may also try to divide it
into several parts, draw one part at a time and connect the adjacent parts with
connecting points. One thus can find out the generative model of digits in one's
own mind. The problem is that a lot of noise is present in the generative
processes. For instance, there can be decorating parts such as serifs, and some
strokes can be exceptionally long or short, when they are not well controlled.

To deal with the noises, we propose to include the singularities as
symbols, and rotate the data pieces for better match. Another thing worth
mentioning is that the symbols are all edges so that the symbols are harder to
match, because of their relatively small domains. 
One way to deal with this problem is to blur the identifying operators around the edges
to give them some breadth. For instance, we blur a circle to make a ring, so that
it can match imperfect circles. The size of the blurring can be added as a new
parameter. 

\section{Conclusion}

In this conceptual paper, we have introduced a Deep Symbolic Networks (DSN) model, which
is inspired by DNN and also comes out from observing the real world - it models
the deep, hierarchical structure of the world, with the observations that humans
symbolize physical matter, and that singularities isolate symbols and create
symbol dictionary naturally for us.

Because of its simple, universal and transparent structure, in addition to the many
advantages of DNN, such as the strong representative ability and the universal learning
methods, it also possesses its own unique strong points. Of course, we need to prove with experiments that it has the same
performance as DNN. Nonetheless, its many unique features are
essential for general artificial intelligence (AI). A deep symbolic network comprises symbols and links,
suitable for Bayesian decision making.
If the symbols and links are complete enough, we conjecture that they can generate human-level intelligence. 
We can check the symbols and links the DSN has learned, and thus assess
what it can do and what it can not. This kind of understanding should allow us to
build more secure AI systems. It is also easier to train, because it is more
friendly to unsupervised learning, it accepts all kinds of data, and it can
accumulate knowledge.

By this conceptual paper, we hope to have presented an idea that will trigger 
an interest to develop it within an open source project, hosted on
\href{https://github.com/qunzhi/Deep-Symbolic-Networks}{Github}, towards the
development of general AI.

\bibliographystyle{plain} 
\bibliography{ref} 
\end{document}